\documentclass[runningheads]{llncs}
\usepackage{graphicx}
\usepackage{cite}
\usepackage{ amssymb }
\usepackage{amsmath}
\usepackage{multirow}
\usepackage{caption}
\usepackage{xcolor}

\begin{document}
%
\title{Progressive DeepSSM: Training Methodology for Image-To-Shape Deep Models}
\titlerunning{Progressive DeepSSM}
%
\author{Abu Zahid Bin Aziz \inst{1,2} \and
Jadie Adams\inst{1,2}
\and
Shireen Elhabian\inst{1,2}
}
\authorrunning{Aziz et al.}
%
\institute{Scientific Computing and Imaging Institute, University of Utah, Salt Lake City, Utah, USA \and
Kahlert School of Computing, University of Utah, Salt Lake City, Utah, USA
\email{\{zahid.aziz,jadie,shireen\}@sci.utah.edu}}
\maketitle              
\begin{abstract}
Statistical shape modeling (SSM) is an enabling quantitative tool to study anatomical shapes in various medical applications.
However, directly using 3D images in these applications still has a long way to go. Recent deep learning methods have paved the way for reducing the substantial preprocessing steps to construct SSMs directly from unsegmented images. Nevertheless, the performance of these models is not up to the mark. 
Inspired by multiscale/multiresolution learning, we propose a new training strategy, progressive DeepSSM, to train image-to-shape deep learning models. The training is performed in multiple scales, and each scale utilizes the output from the previous scale. This strategy enables the model to learn coarse shape features in the first scales and gradually learn detailed fine shape features in the later scales. 
We leverage shape priors via segmentation-guided multi-task learning and employ deep supervision loss to ensure learning at each scale. 
Experiments show the superiority of models trained by the proposed strategy from both quantitative and qualitative perspectives. This training methodology can be employed to improve the stability and accuracy of any deep learning method for inferring statistical representations of anatomies from medical images and can be adopted by existing deep learning methods to improve model accuracy and training stability. 


\end{abstract}
\section{Introduction}
\label{sect:introduction}
Statistical shape modeling (SSM) has become vital for quantitative studies of biological and medical data by providing a statistically consistent geometrical description for each shape across a given population. Recent progress in this field has enabled a wide range of clinical and scientific SSM applications, such as bone reconstruction in orthopedics from 2D or 3D medical images \cite{fuessinger2019virtual, harris2013statistical}, atrial fibrillation in cardiology\cite{bhalodia2018deep, gardner2013point}, brain ventricle analysis in neuroscience \cite{gerig2001shape, zhao2008hippocampus, biffi2020explainable}. 

Several shape representations have been introduced and utilized. Among them, deformation-based and correspondence-based models are the most popular \cite{beg2005computing, cates2007shape}. While deformation fields can represent shapes directly from images, in this work, we have opted for correspondence-based shape representation as it does not require a reference/atlas. 
Nevertheless, the proposed training strategy can also be adapted to deformation-based shape representation. Correspondence-based models (also known as point distribution models or PDMs) utilize an ordered set of landmarks or correspondence points placed on the shape surface in a consistent manner across the population. Several algorithms are available for these types of shape representation \cite{davies2002minimum, styner2006framework,cates2007shape}. Each algorithm follows a set of time-consuming and labor-intensive preprocessing steps which require domain expertise, including shape segmentation, resampling, smoothing, and alignment. Furthermore, PDM optimization processes and inference on new shapes are computationally expensive and time-consuming.

To ease the burden of manpower and heavy-duty preprocessing, deep learning-based models have been proposed to harness the power of data to learn a functional mapping directly from images to statistical representations of shapes \cite{bhalodia2018deepssm, bhalodia2021deepssm, adams2020uncertain, adams2022images, adams2023fully}. These works provide a considerable advantage over conventional PDM methods in inference, as they do not require prohibitive, manual preprocessing steps and computationally complex re-optimization. Once a deep network is trained, a PDM can be inferred from a new unsegmented image in seconds. However, in terms of accuracy performance, existing deep learning models have yet to be up to the mark. Here, we propose a training strategy based on progressive learning, deep supervision, and multi-task learning to improve the performance of existing deep learning models. 



The proposed methodology draws inspiration from three key concepts: progressive learning that builds on knowledge from prior learned tasks \cite{karras2017progressive}, deep supervision that applies loss to intermediate neural network layers \cite{wang2015training}, and multi-task learning \cite{zhou2021multi} that leverages commonalities and differences across related tasks to improve generalization. The model consists of several progressive blocks, and each block is trained to predict an increasing number of correspondence points, i.e., a shape descriptor or representation at a specific scale. In other words, we predict the correspondence points in a multiscale training process, where each scale leverages the previous scales' output to predict the points. 
We provide a thorough architecture investigation, exploring the advantages and disadvantages of shared block backbones and the inclusion of an auxiliary segmentation task for improved PDM prediction. 
Furthermore, we have employed deep supervision to train our models and explored three loss calculation strategies that depend on the intermediate layers where the loss is applied. Finally, we demonstrate that the proposed training strategies significantly improved performance during the training and testing of the existing models. These training strategies can provide effective deep learning-based PDMs for accurate shape representation from images.

\section{Related Works}
\label{sect:related_works}

We discuss the related works from three points of view: deep learning-based SSMs, progressive learning, and deep supervision.

\textbf{Deep learning-based SSM methods:} DeepSSM is a state-of-the-art model that can provide statistical shape representations directly from images \cite{bhalodia2018deepssm, bhalodia2021deepssm}. It uses a principal component analysis (PCA) based data augmentation scheme and has achieved good results on downstream tasks \cite{bhalodia2018deep}. Probabilistic variants of DeepSSM that add uncertainty quantification have been proposed. For example, Uncertain-DeepSSM focuses on predicting data-dependent and model-dependent uncertainties to overcome the overconfident estimation of the deep learning models \cite{adams2020uncertain}. Recently, VIB-DeepSSM and it's fully Bayesian extension have been proposed, which utilize variational information bottleneck to capture the latent representation rather than regressing PCA scores \cite{adams2022images}. 
All of these works supervise the entire dense set of correspondence points compared to the iterative process of incrementally predicting correspondence points of the conventional PDMs.  This single-step regression process is error-prone in complex shape regions. The proposed training method can be used in addition to any of these methodologies to achieve more stable training and better performance. 

\textbf{Progressive learning:} Since its introduction in 2017, progressive learning \cite{karras2017progressive} has revolutionized the training process for generative adversarial networks (GAN) and learning applications such as shape representation \cite{park2019deepsdf}, speech recognition \cite{fayek2020progressive, gao2016snr}, and person re-identification \cite{wu2019progressive}. This incremental training process allows the model to learn a high-level, coarse output representation first, then gradually move on to detailed low-level, fine features. In the context of our task, rather than mapping the feature vectors directly to the final number of correspondence points, progressive learning allows us to map it to a lower number of points first, then gradually increase it to the final number to provide a better shape representation. Here, the mapped points in each scale cover the whole shape.

\textbf{Deep Supervision:} Deep supervision has improved training performances by adding losses in intermediate network layers in a wide range of applications, such as edge detection \cite{liu2016learning}, image segmentation \cite{zhang2018deep}, 2D/3D keypoint localization \cite{li2018deep}, and image classification \cite{wang2015training, li2018deep}. We leverage this approach by adding supervision in each level of correspondence point prediction, allowing the model to converge better than existing methods. 

 
\section{Methodology}
\label{sect:methodology}


\subsection{Datasets}
\label{sub:dataset}
We showcase the proposed training method using two datasets: femur and left atrium. 


\noindent\textbf{Femur dataset:} The femur dataset comprises 59 CT scans, with 49 identified as control scans, showcasing healthy subjects without any morphological irregularities in the femur bone. The remaining 10 scans are diagnosed with CAM-FAI, which is a morphological abnormality of the femur characterized by a lack of normal concavity at the femoral head-neck junction \cite{harris2013statistical}. From this pool, we randomly incorporate 42 control images and 8 CAM-FAI images into the training set, reserving the remaining images for testing. Image downsampling is performed at a rate of 50\%, resulting in an image size of $130 \times 92 \times 117$ and maintaining a uniform voxel spacing of 1.0 mm.

\noindent\textbf{Left atrium dataset:} The left atrium dataset encompasses 206 late gadolinium enhancement (LGE) MRI images from patients diagnosed with atrial fibrillation (AF), which results in irregular heart rhythm due to abnormal electrical impulses firing in the atrium. Similar to the femur data processing, we downsample these images by 50\%, reaching a resolution of $118 \times 69 \times 88$ with a uniform voxel spacing of 1.25 mm. We randomly split the instances into 176 images for training, leaving 30 images for the testing phase.

\subsection{Training Data}

In constructing the multi-scale training data, we first determine the desired number of scales for the progressive training architectures based on the number of correspondence points in the first and last scales. The initial number of points is set at 256 to ensure enough coverage to capture coarse shape features. The maximum number of correspondence points (1024) for specific anatomy is selected empirically, as per the anatomy's size, curvature, and morphological variations. This process is executed using ShapeWorks \cite{cates2017shapeworks} coarse-to-fine particle splitting strategy until the final correspondence points representation adequately captured the given anatomy's detail. We have selected ShapeWorks to generate PDMs at each scale (256, 512, 1024) because of its ability to generate PDMs with consistent qualitative and quantitative performance \cite{goparaju2022benchmarking}. The ground truth PDMs for the test dataset are generated using the pre-optimized shape models of the training data.

Due to the low-sample size that is typical in medical imaging, we have applied model-based data augmentation \cite{bhalodia2021deepssm,adams2020uncertain} to generate additional realistic training examples. To do so, we applied principal component analysis (PCA) at each level of correspondence point density. A set of $M$ 3D correspondence points for $N$ samples, denoted by $\bigl\{y_{n}\bigr\}_{n=1}^N$ where $y_n \in \mathbb{R}^{3 M}$, is reduced to $\mathrm{L}$ dimensional PCA scores $z_n \in \mathbb{R}^L$ where $\mathrm{L}$ is relatively low (between 15 and 25). These PCA scores can be expressed by a mean vector $\mu \in \mathbb{R}^{3 M}$, a diagonal matrix of eigenvalues $\lambda \in \mathbb{R}^{L \times L}$ and matrix of eigenvectors $v \in \mathbb{R}^{3 M \times L}$ by the equation: $z_n=v^T\left(y_n-\mu\right)$. A distribution is fit to the PCA scores via kernel density estimation (KDE). For the femur data, we use 20 PCA modes (which captured around 99\% of the population variability), and for the left atrium, we used 25 PCA modes (which captured around 97\%  of the variability).To generate a synthetic image, we first draw a random sample from the KDE distribution and then use a technique that involves finding the closest example from a set of input images.
This is the same augmentation technique as DeepSSM, and more details can be found there \cite{bhalodia2021deepssm}. 

We have generated 5000 augmented image/correspondence point pairs for the femur dataset and 4000 for the left atrium dataset. The augmented and original images are used for training and validation in an 80:20 ratio. 

\begin{figure}[t]
\centering
\includegraphics[width=12cm, height=7cm]{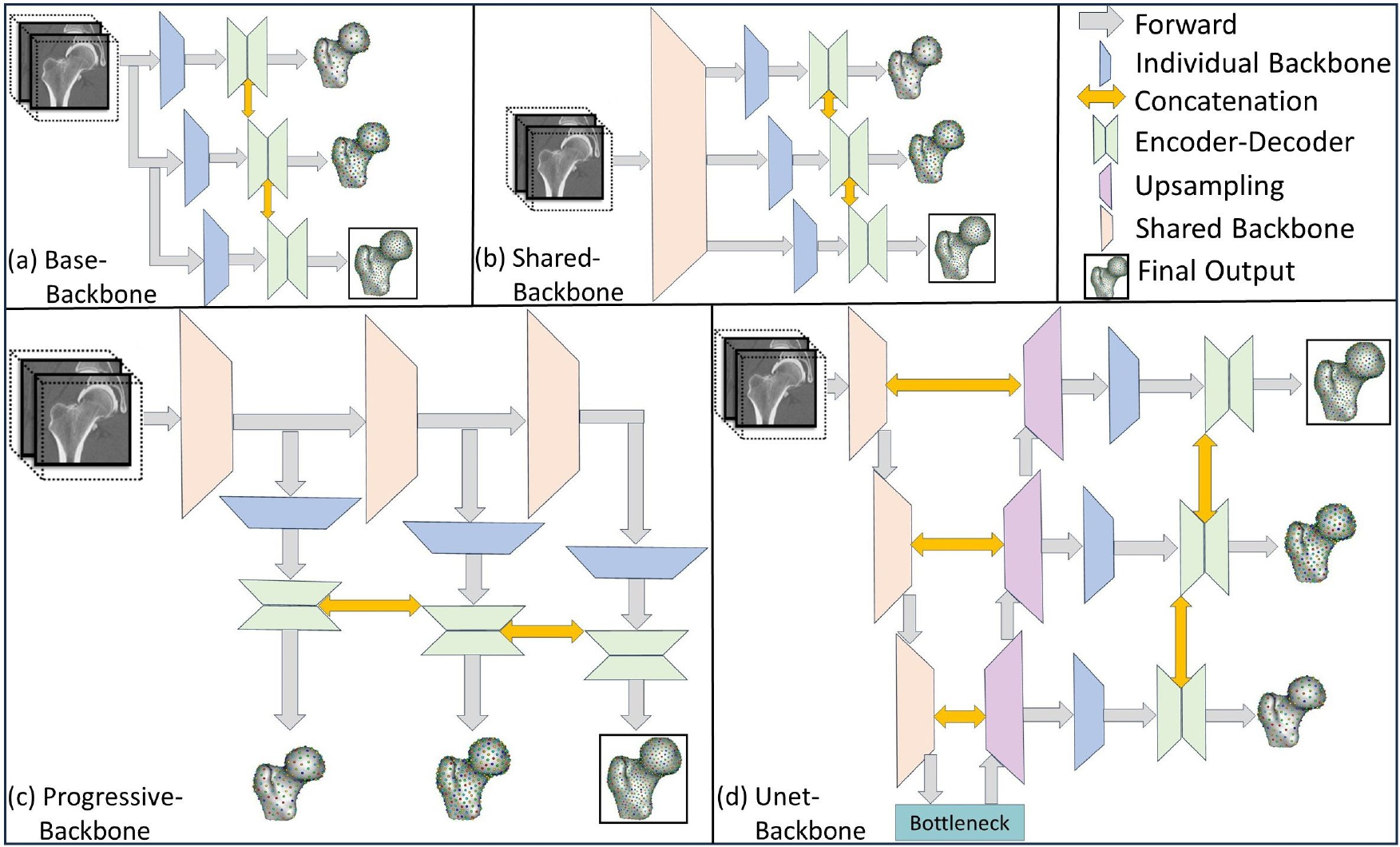}
\caption{Proposed model architecture with (a) Base-Backbone, (b) Shared-Backbone, (c) Progressive-Backbone and (d) Unet-Backbone. 
} \label{fig:model}
\end{figure}

\subsection{Model Architecture}
\label{sub:architecture}

In this work, we have adopted DeepSSM \cite{bhalodia2018deepssm} as 
the primary building block to underscore the impact of the training strategy. DeepSSM and its various offshoots use a single deep network to estimate the complete set of correspondence points at the highest resolution/scale directly from unsegmented images. This work aims to demonstrate the efficacy of a multi-scale, progressive learning strategy when used to train these models. Although we demonstrate the proposed work using the original DeepSSM network, the proposed training and loss strategies can be readily applied to other variants. The presented investigation entails four variants of architecture, each contingent on the backbone of every scale. 



\begin{itemize}
    \item \textbf{Base-Backbone}: To obtain evidence of the progressive training's improved performance, we need to conduct a proof of concept. Hence, we have experimented with the base progressive architecture (Figure \ref{fig:model}(a)), whereby each scale is predicted by an individual DeepSSM block. From the second scale onwards, every block incorporates latent features from preceding blocks as an auxiliary input. This base architecture, applied initially and subsequently, utilizes a Convolutional Neural Network (CNN) backbone, comprising five convolutional and three max-pooling layers. Following each backbone, an encoder-decoder network is deployed. The encoder is built from three fully-connected layers, with the final layer containing the same number of nodes as the number of PCA modes ($\mathrm{L}$). The decoder comprises a single layer with $\mathrm{3M}$ nodes, where M corresponds to the number of correspondence points for that particular scale. The decoder is initialized with the eigenvalues $(\left.v z_n+\mu\right)$ derived from the principal components serving as weights and the mean shape acting as bias. 

    \item \textbf{Shared-Backbone}: Once we have the proof of concept, our goal evolves to explore whether all scales share predictive image features for shape features at different scales, or if each scale requires its own feature extraction. To achieve this, we have implemented an approach of dividing the backbone into two parts: one common for all blocks and the other distinctive for each block. Specifically, we utilize two convolutional and one max-pooling layer from the backbone of the base architecture as the shared backbone, while the remaining three convolution and two map-pooling layers are used individually for each block. The encoder-decoder network is the same as the Base-Backbone architecture. This architecture is shown in Figure \ref{fig:model}(b).

    \item \textbf{Progressive-Backbone}: The previous network utilizes an identical architecture for all scales. However, we are curious to explore the potential benefits of incorporating more layers as the number of correspondence points rises with each scale. Therefore, we have conducted experiments with a progressive backbone, where the number of layers in the shared backbone increases as we progress to later scales (Figure \ref{fig:model}(c)). Despite these changes, the encoder-decoder architecture remains consistent.
    
    \item \textbf{Unet-Backbone}: Along with weight sharing between the blocks, we want to explore multitasking capabilities  and investigate the performance of these models. Specifically, we focus on the task of segmentation, which is a fundamental prerequisite for non deep learning based methods. To achieve this, we have integrated the Progressive-Backbone model into the popular U-net segmentation architecture and pass a fusion of the bottleneck and decoder features to predict each scale of correspondence points (Figure \ref{fig:model}(d)) \cite{ronneberger2015u}. Motivation for fusing the bottleneck and decoder features is provided in section \ref{sect:ablation}. The segmentation-guided backbone provides the network with a shape prior, increasing correspondence prediction accuracy. The fused bottleneck and decoder features are passed to a feature extractor consisting of convolution and max-pooling layers to acquire the feature space for each scale. The feature space is connected to an encoder-decoder network, similar to previous architectures, to predict their respective correspondence points. However, in this case, we have not initialized the decoder using PCA, as this imposes linearity on the shape, which may hurt accuracy in the case of complicated shapes.
    
\end{itemize}

Each of these architectures uses  a filter size of 5 for the convolutional layers and 2 for the max-pooling layers, which are chosen empirically. To ensure optimal performance, we have applied batch normalization after each convolution operation, followed by parametric ReLU (PReLU) activation. 

\subsection{Loss Function}
\label{sub:loss}


Our model employs mean squared error (MSE) loss for the predicted  correspondence points for each scale. For the ground truth $y_k$ and predictions $\hat{y}_k$ for any scale k and N number of samples, the MSE loss is defined as:

\begin{equation}
    \mathcal{L}_k = \frac{1}{N} \sum_{n=1}^{N} (y^k_n - \hat{y}^k_n)^2
\end{equation}

We have employed three loss variants in our methodology. They are defined as follows:
\begin{itemize}
    \item \textbf{Fixed}:   For any scale k,  the previous scales' weights are frozen and the total loss is defined as:
    \begin{equation}
    \mathcal{L}_{Fixed} = \mathcal{L}_k
    \end{equation}
    \item \textbf{Shallow-Supervision}: For any scale k, the total loss is defined as the sum of the MSE loss of that scale and the previous scale. Here, during the training of each scale, the loss is backpropagated until the previous scale.
    \begin{equation}
   \mathcal{L}_{Shallow-Supervision}= 
    \begin{cases}
        \mathcal{L}_k + \mathcal{L}_{k-1}, & \text{if }  k > 0\\
        \mathcal{L}_k,              & k=0
    \end{cases}
    \end{equation}
    
    \item \textbf{Deep-Supervision}: For any scale k, the total loss is defined as the summation of the MSE loss of the initial scale to that scale.
    \begin{equation}
    \mathcal{L}_{Deep-Supervision} =  \sum_{i=0}^{k} \mathcal{L}_i
    \end{equation}
\end{itemize}

In addition to the aforementioned correspondence loss, we have incorporated a segmentation loss for the Unet-Backbone models. 
Consequently, the cumulative loss is quantified using the subsequent formula:

\begin{equation}
\mathcal{L}_{total} = \alpha * \mathcal{L}_{seg} + (1 - \alpha) * \mathcal{L}_{PDM}
\end{equation}

In this context, $\alpha$ represents an empirically determined hyperparameter designed to balance the weights between the segmentation and correspondence losses. $\mathcal{L}_{seg}$ corresponds to the binary cross-entropy (BCE) loss between the original and the predicted segmentation, and $\mathcal{L}_{PDM}$ refers to any one of the aforementioned loss variants (Fixed, Shallow-Supervision, Deep-Supervision).


\subsection{Evaluation Metric}
\label{sub:metric}


We use two key metrics to evaluate the effectiveness of the proposed methodology: Root Mean Square Error (RMSE) and surface-to-surface distance (in mm). 
RMSE is calculated as the square root of the average squared differences between the predicted and actual observations. Specifically, we average the RMSE for the x, y, and z coordinates, where N is the total number of 3D correspondences:


\begin{equation}
    RMSE =\frac{1}{3}\left(R M S E_x+R M S E_y+R M S E_z\right)
\end{equation}

Where,  for N sets of ground truth and predicted correspondence points at scale k, $R M S E_x=\sqrt{\frac{\left|y_{nx}^k-\hat{y}_{nx}^k\right|_2^2}{N}}$ and similar for y and z coordinates.

Surface-to-surface distance is computed by converting the ground truth and predicted points to meshes and calculating the euclidean distance from each vertex of the ground truth mesh to the closest face of the predicted mesh. The reported values are the average of the vertex-wise surface-to-surface distance from the ground truth to the predicted shapes.

\subsection{Training Procedure}
\label{sect:training}


We have employed a multiscale, progressive training strategy to ensure better convergence. This means we train one scale at a time, and only after that scale reaches convergence do we move on to the next scale. We have used a Cosine Annealing learning rate scheduler \cite{loshchilov2016sgdr} to update each epoch's learning rate. The rapid change in the learning rate of this scheduler has helped to make sure the learning process is not stuck at a local minimum during training. The initial learning rate is set to 0.001, and Adam optimization is used. Each scale is trained for a maximum of 50 epochs with a batch size of six. However, to avoid overfitting, we have employed an early stopping strategy, where we stop the training if the validation loss is not improved after 15 consecutive epochs.

In the case of the Unet-Backbone models, the segmentation component is trained first for five epochs to ensure a good shape prior to the correspondence prediction. Then each scale is trained as previously explained. The value of the $\alpha$ parameter for the $\mathcal{L}_{total}$ is empirically set to 0.1.

The training process is implemented in PyTorch, and training is performed on a 12th Gen Intel(R) Core(TM) i9-12900K Desktop with 128 GB RAM and NVIDIA RTX A5000 GPU.

\section{Results}
\label{sect:results}

\begin{figure}[t]
\includegraphics[width=\textwidth]{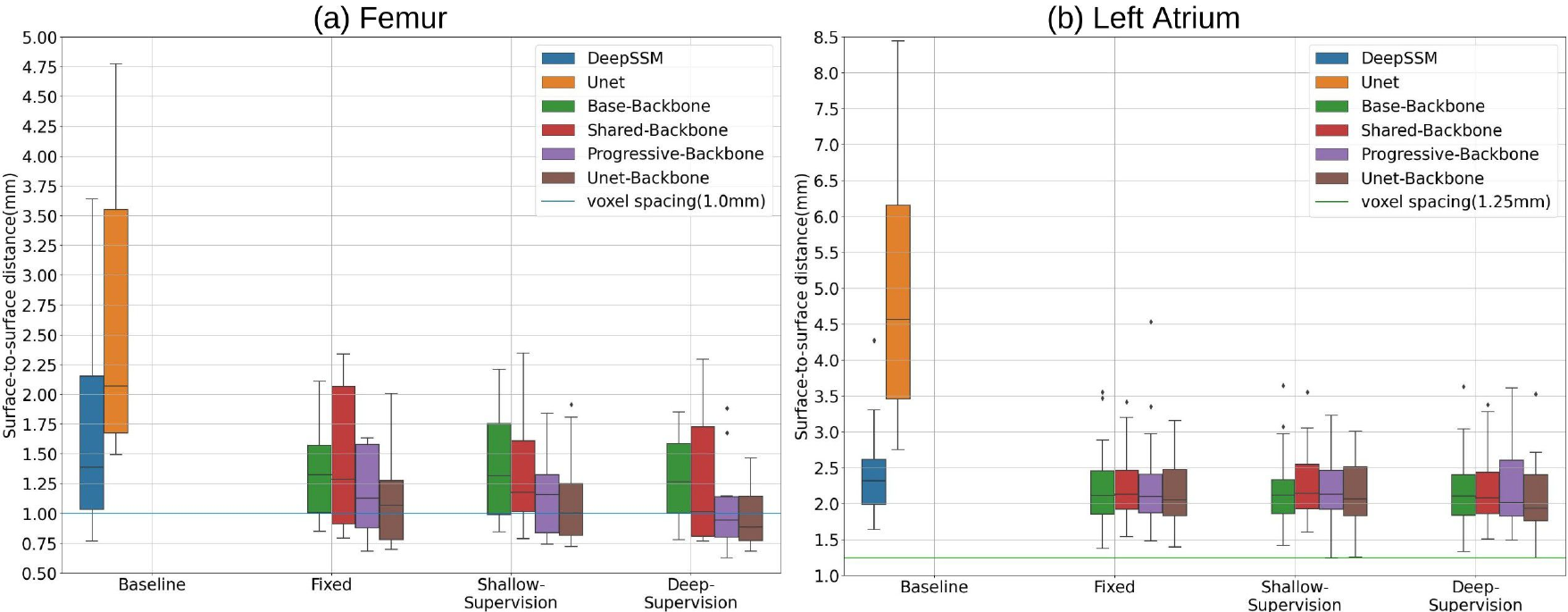}
\caption{Surface-to-surface distance comparison of our proposed models with DeepSSM(\cite{bhalodia2021deepssm}) in (a) femur and (b)left atrium dataset.  The black line in each boxplot marks the median value, and the blue horizontal line represents the voxel spacing of the images.} \label{fig:femur_box}
\end{figure}

\subsection{Femur}
\label{sect:femur}
We have trained each model for three different loss functions as described in section \ref{sub:loss}. The surface-to-surface distance comparison is shown in a boxplot in Figure \ref{fig:femur_box}(a). The y-axis shows the two baselines (DeepSSM \cite{bhalodia2021deepssm} and Unet for segmentation) and the three loss variants. For each loss variant column, different boxplots denote different model architectures. The blue and orange boxplot represents the baseline DeepSSM and Unet results, whereas the green, red, purple, and brown boxplots represent the Base-Backbone, Shared-Backbone, Progressive-Backbone, and Unet-Backbone, respectively. 

We observe a consistent trend in Figure \ref{fig:femur_box}(a) across all model architectures, namely that deep supervision enables the model to make more accurate predictions of correspondence points, resulting in more accurate shapes. This suggests that the progressive training strategy is benefitting from the deep supervision, as the gradients from the later scales, which capture fine-scale shape features, are used to fine-tune the earlier scales. This allows for improved conditions for the input signal for the finer scales, as the scales are not independent; each training iteration contributes to learning the fine-shape features.

The model performance remains consistently high across various architectures. Generally, the Base-Backbone models have shown slightly better results than the baseline, indicating that progressive architectures can yield improved outcomes. Furthermore, the Shared-Backbone and Progressive-Backbone models outperform the Base-Backbone. Notably, multitasking with a progressive backbone proves most effective in enhancing performance, as evidenced by the results of the Unet-Backbone. We have also sought to compare our approach with standard segmentation architecture, which calculates the surface-to-surface distance between original and predicted segmentations. The proposed SSM models fared much better than segmentation-based models in reconstructing shapes from images. 

\begin{table}[t]
\centering
\caption{The comparison between DeepSSM and the proposed models on the Deep Supervision loss in test data in terms of RMSE.}\label{tab1}
\begin{tabular}{|c|ccccc|}
\hline
\multirow{2}{*}{\textbf{Dataset}} & \multicolumn{5}{c|}{\textbf{RMSE on Test Data (Mean $\pm$ Standard Deviation)}}   \\ \cline{2-6} 
              & \multicolumn{1}{c|}{\textbf{DeepSSM}} & \multicolumn{1}{c|}{\textbf{\begin{tabular}[c]{@{}c@{}}Base-\\ Backbone\end{tabular}}} & \multicolumn{1}{c|}{\textbf{\begin{tabular}[c]{@{}c@{}}Shared-\\ Backbone\end{tabular}}} & \multicolumn{1}{c|}{\textbf{\begin{tabular}[c]{@{}c@{}}Progressive-\\ Backbone\end{tabular}}} & \textbf{\begin{tabular}[c]{@{}c@{}}Unet-\\ Backbone\end{tabular}} \\ \hline
\textbf{Femur} & \multicolumn{1}{c|}{$1.37 \pm 0.72$}      & \multicolumn{1}{c|}{$1.15 \pm 0.52$} & \multicolumn{1}{c|}{$1.07 \pm 0.41$} & \multicolumn{1}{c|}{$0.93 \pm 0.34$} & \textbf{$0.78 \pm 0.21$}   \\ \hline
\textbf{Left Atrium}             & \multicolumn{1}{c|}{$1.72 \pm 0.8$}       & \multicolumn{1}{c|}{$1.65 \pm 0.42$}                                  & \multicolumn{1}{c|}{$1.62 \pm 0.45$}                                                                          & \multicolumn{1}{c|}{$1.55 \pm 0.43$}                                                                               & \textbf{$1.48 \pm 0.28$}   \\ \hline
\end{tabular}
\end{table}

We have compared the proposed models with the DeepSSM in terms of RMSE (Table \ref{tab1}) which, unlike surface-to-surface distance, captures whether or not the points are in correspondence. We can see a significant improvement in RMSE error for the Progressive-Backbone (32.12\%) and Unet-Backbone models (43.06\%). This improvement shows the superiority of the proposed models in the test data. 

\begin{figure}[t]
\centering
\includegraphics[width=8cm, height=5.5cm]{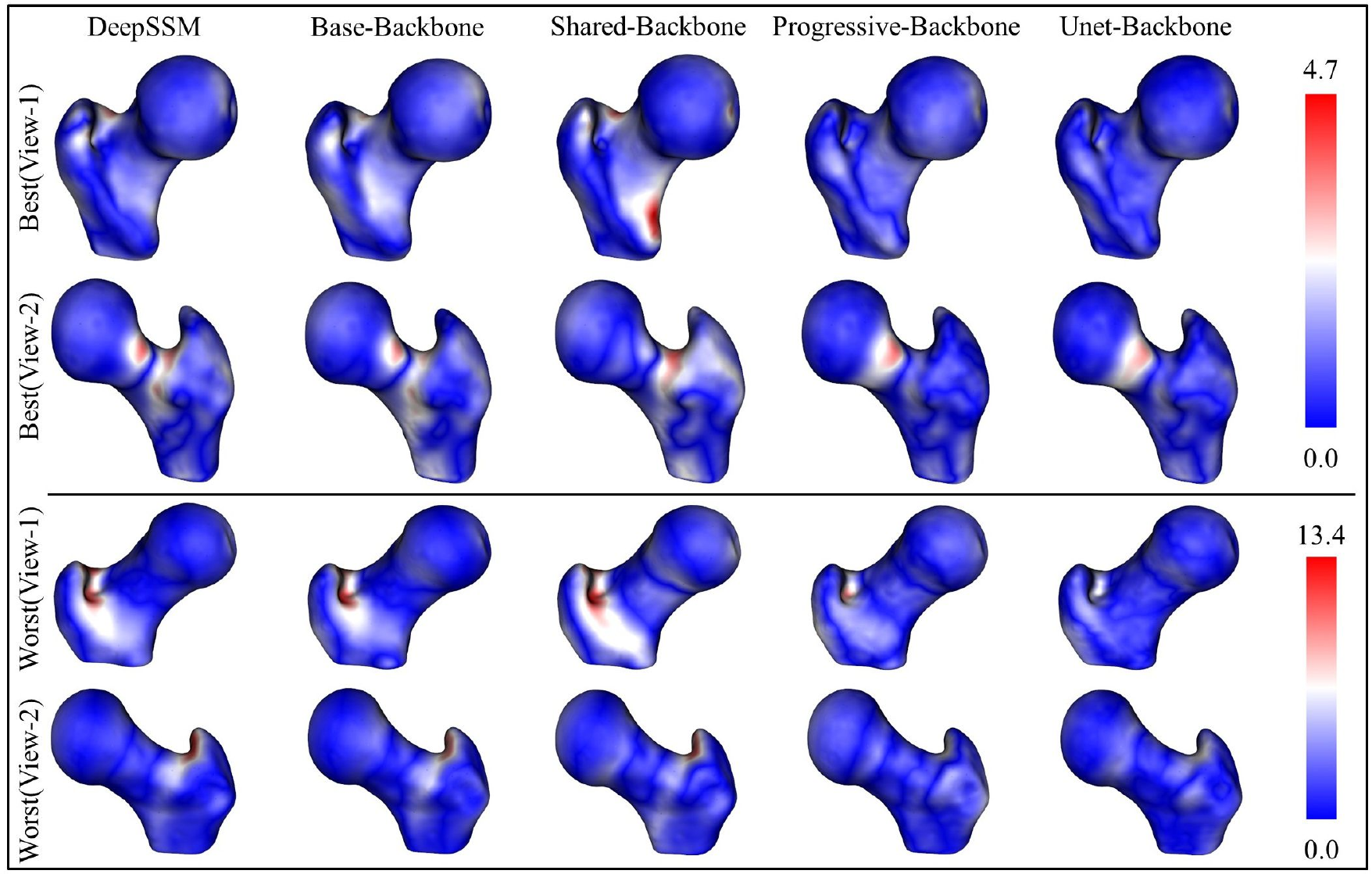}
\caption{Reconstruction error of the models’ output is shown as a heatmap on the ground truth meshes for DeepSSM’s best and worst output in the test data. The models reported in the figure are trained with Deep-Supervision losses} \label{fig:femur_cases}
\end{figure}

Additionally, we quantitatively evaluate the performance of our proposed models for 3D mesh reconstruction by comparing the reconstruction errors via heatmaps on the ground truth meshes. Specifically, we select the best and worst outputs of DeepSSM on the test data based on surface-to-surface distance and compare them with the proposed models' predictions. From this analysis, we generate error maps for our models' prediction on the ground truth mesh for the selected samples. The results of the comparison are shown in Figure \ref{fig:femur_cases}. Our findings show a significant improvement in the proposed models' prediction, particularly for the Progressive-Backbone and Unet-Backbone models for both cases.

\begin{figure}[t]
\centering
\includegraphics[width=10cm, height=4cm]{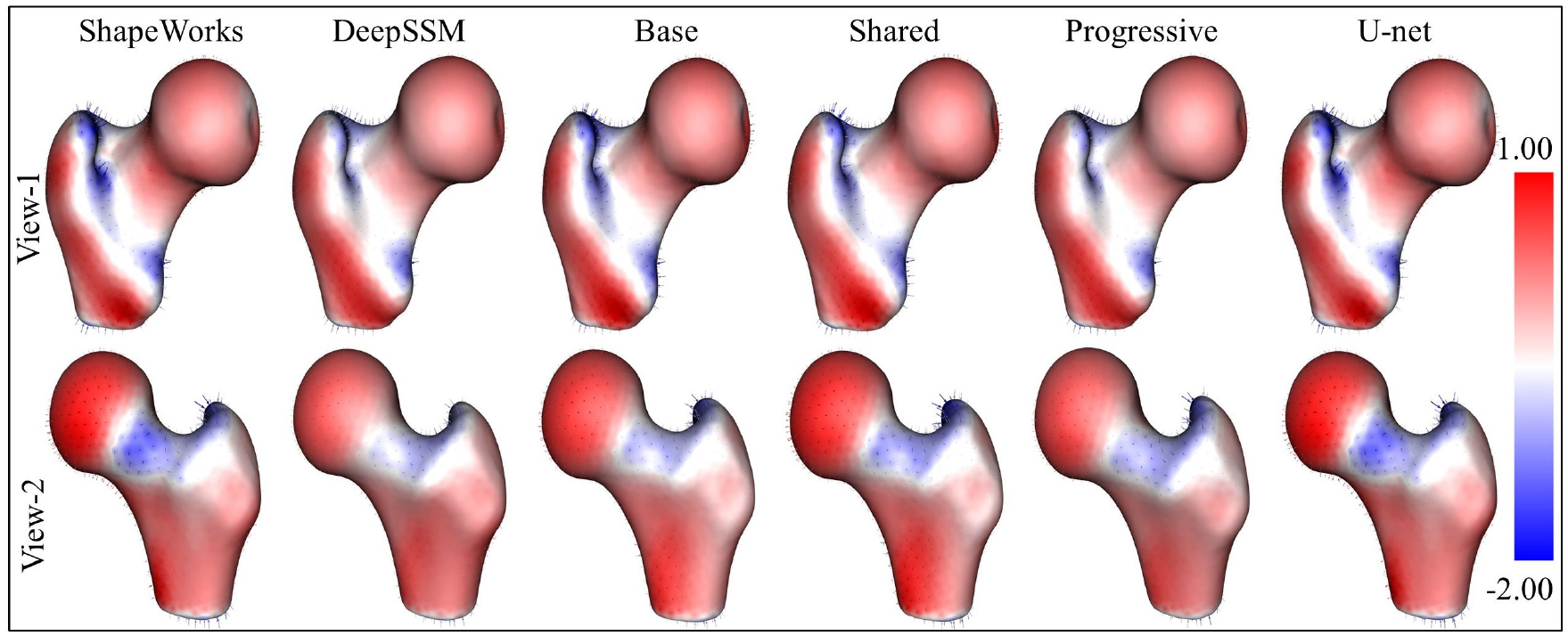}
\caption{Group Difference comparison of our proposed models with the original DeepSSM\cite{bhalodia2021deepssm} and ShapeWorks\cite{cates2017shapeworks}.} \label{fig:group_diff}
\end{figure}

\textbf{Downstream Task - Group Differences:}  It is clinically significant to capture the statistical morphological difference between the CAM-FAI shape and the typical femur bone shape. In this experiment, we have employed models trained with Deep-Supervision loss. Our approach involves the construction of two groups - one for controls and one for pathology (CAM-FAI) - and computing the difference between their means ($\mu_{normal}$ and $\mu_{cam}$). By doing so, we were able to showcase this difference on a mesh, which is known as group difference \cite{harris2013statistical}. To achieve this, we utilize the ShapeWorks' PDM model, DeepSSM and the proposed models' predicted particles, using the entire data for testing and training. Our findings demonstrate a strong similarity between the ShapeWorks and the proposed models' group differences, particularly for the Progressive-Backbone and Unet-backbone models (Figure \ref{fig:group_diff}). This suggests that the proposed models can effectively obtain correspondences without the need for heavy pre-processing and segmentation steps. This ability to characterize the CAM deformity is crucial in observing the expected outcome of femur anatomy smoothly exhibiting inward motion around the CAM lesion as observed in clinical practice. Our results indicate that the proposed models have the potential to be a valuable tool in the analysis of femur anatomy and can aid in the diagnosis and treatment of CAM deformities.

\subsection{Left atrium}
\label{sect:la}

The left atrium MRI dataset presents significant variations in intensity and quality, influenced by the topological differences related to the arrangements of pulmonary veins. Similar to the femur dataset, we have trained our models for all loss variants explained in \ref{sub:loss}. The results are shown in Figure \ref{fig:femur_box}(b). We can see that the Deep-Supervision loss helps in the case of this dataset as the surface-to-surface distance is much better for this loss compared to the Fixed and Shallow-Supervision. 

All of the models proposed outperformed the baseline results. Interestingly, the SSM-based methods seem to be generating better 3D shapes compared to the standard segmentation baseline. The high variability of the dataset is likely contributing to the Unet model's underwhelming performance. Regarding different model architectures, the proposed models have demonstrated similar performance for a specific loss type, with Unet-Backbone slightly edging ahead. The trend in performance is consistent with the femur dataset, where the Progressive-Backbone and Unet-Backbone models surpass the Base-Progressive and Shared-Progressive models. By examining Table 1, it is apparent that the suggested models have shown significant improvement, particularly the Progressive-Backbone (9.88\%) and Unet-Backbone (13.95\%) models concerning RMSE in the test data. These enhancements in both evaluation metrics highlight the advantageous impact of the proposed training techniques, especially when dealing with complex datasets.

\begin{figure}[t]
\centering
\includegraphics[width=10.2cm, height=6cm]{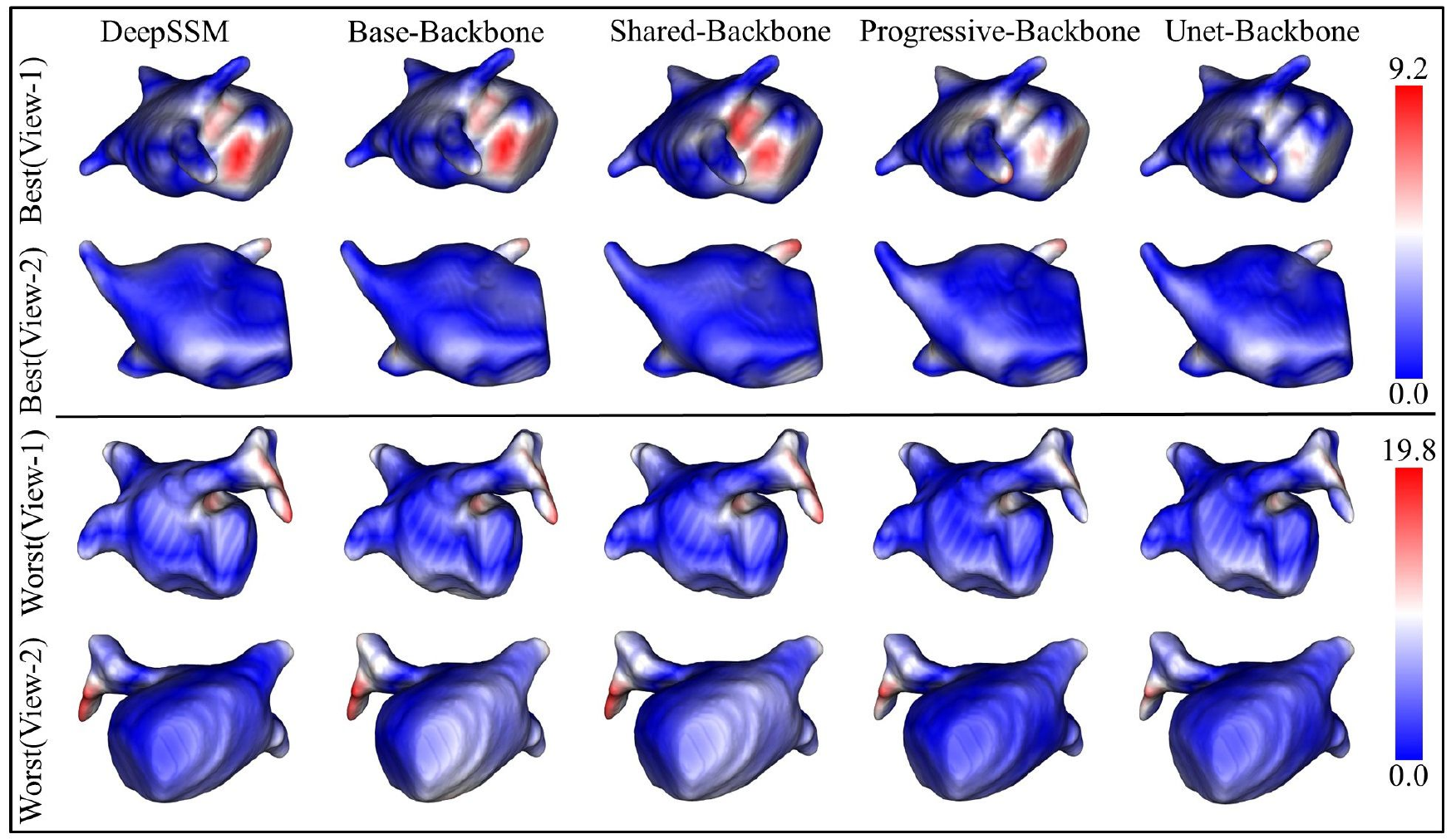}
\caption{Reconstruction error of the models’ output is shown as a heatmap on the ground truth meshes for DeepSSM’s best and worst output in the test data} \label{fig:la_cases}
\end{figure}

Furthermore, we have conducted a thorough analysis of the reconstruction error comparison for the best and worst output of DeepSSM in the test data with respect to surface-to-surface distance. Our findings indicate that the output for both the best and worst case of the proposed models have significantly enhanced DeepSSM's outputs, as illustrated in Figure \ref{fig:la_cases}. This outcome is consistent with the femur dataset, which further validates the effectiveness of the proposed models in improving the accuracy of surface reconstruction. 

\textbf{Downstream Task - Atrial Fibrillation Recurrence Prediction:} Atrial Fibrillation (AF) is a medical condition characterized by an irregular heartbeat. To treat AF, doctors often use a therapeutic procedure called catheter ablation. Unfortunately, some patients may experience a recurrence of AF even after undergoing ablation. The left atrium dataset includes binary labels for each patient indicating whether they had AF recurrence following ablation. 

To train a multi-layer perceptron (MLP) model for classifying AF recurrence, we utilized PCA scores from ground truth data and the latent features of the encoder-decoder network for both the DeepSSM and proposed models. Our experiments employ the same training, validation, and test sets, and we use the Deep-Supervision trained models. The results, presented in Table \ref{tab2}, demonstrate a significant performance improvement compared to the DeepSSM. Notably, our Unet-Backbone model even outperforms the ShapeWorks accuracy. We believe that the Unet-Backbone's encoding of image-based features, not available in PDM, contributes to its success in this downstream task.

\begin{table}[t]
\centering
\caption{The comparison between baselines and the proposed models' accuracy on AF recurrence.}\label{tab2}
\begin{tabular}{|c|c|c|c|c|c|c|}
\hline
\textbf{Metric(\%)} & \textbf{ShapeWorks} & \textbf{DeepSSM} & \textbf{\begin{tabular}[c]{@{}c@{}}Base-\\ Backbone\end{tabular}} & \textbf{\begin{tabular}[c]{@{}c@{}}Shared-\\ Backbone\end{tabular}} & \textbf{\begin{tabular}[c]{@{}c@{}}Progressive-\\ Backbone\end{tabular}} & \textbf{\begin{tabular}[c]{@{}c@{}}Unet-\\ Backbone\end{tabular}} \\ \hline
Accuracy    & 63.33        & 56.66            & 58.7         & 60.0    & 61.66    & \textbf{73.33}                                                    \\ \hline
\end{tabular}
\end{table}

\section{Ablation Studies}
\label{sect:ablation}

\begin{figure}[ht]
\centering
\includegraphics[width=5cm, height=4.5cm]{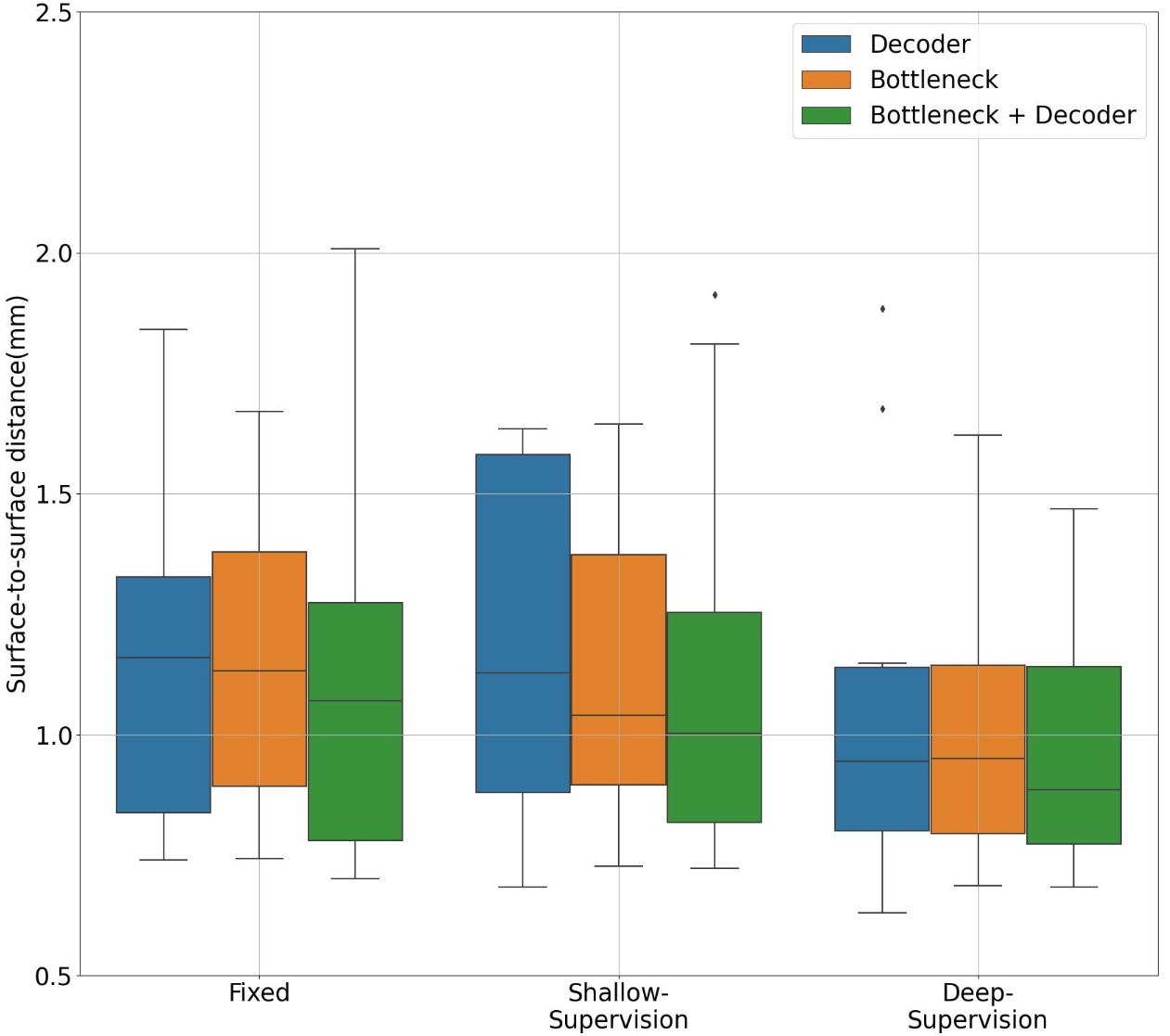}
\caption{Surface-to-surface distance comparison of different features in femur dataset.} \label{fig:ablation_feature}
\end{figure}

We have conducted an ablation study to analyze the impact of different components within our Unet-Backbone architecture. The study focuses on three key areas: the decoder, the bottleneck, and a fusion of the bottleneck and decoder features. The decoder plays a crucial role in reconstructing the spatial information lost during the encoding process. However, when used alone, it may not capture complex object details due to the lack of context. In contrast, the bottleneck condenses the input image information into a more manageable form, making it easier to extract high-level features. However, relying only on the bottleneck for feature extraction may result in a loss of information, especially for larger and more complex inputs. Our study shows that the fusion of bottleneck and decoder features produces the best results in the Unet-Backbone models (Figure: \ref{fig:ablation_feature}). This approach combines the strengths of the decoder and the bottleneck, resulting in a more robust representation.


\section{Conclusion}
\label{conclusion}
Performing statistical shape modeling directly on images is a difficult task. Many image quality complications, such as artifacts, spatial resolution, signal-to-image ratio, etc., make it challenging to perform shape modeling. Hence, our work proposes a multiscale training methodology to learn the features gradually. The proposed training method utilizes multi-tasking based progressive learning and deep supervision to provide better performance. We have tested our methodology on two different datasets with different types of images (CT and MRI scans), and the proposed models provide improved results in both cases. This training method can be integrated into any deep learning-based shape models and achieve better performance. These contributions will help accelerate the adoption of automated statistical shape modeling from images in clinical use cases.

\section*{Acknowledgements}
The National Institutes of Health supported this work under grant numbers NIBIB-U24EB029011, NIAMS-R01AR076120, and NIBIB-R01EB016701. The content is solely the responsibility of the authors and does not necessarily represent the official views of the National Institutes of Health.
%
%
%
\bibliographystyle{splncs04}
\bibliography{samplepaper}

\begin{thebibliography}{10}
\providecommand{\url}[1]{\texttt{#1}}
\providecommand{\urlprefix}{URL }
\providecommand{\doi}[1]{https://doi.org/#1}

\bibitem{adams2020uncertain}
Adams, J., Bhalodia, R., Elhabian, S.: Uncertain-deepssm: From images to probabilistic shape models. In: International Workshop on Shape in Medical Imaging. pp. 57--72. Springer (2020)

\bibitem{adams2022images}
Adams, J., Elhabian, S.: From images to probabilistic anatomical shapes: A deep variational bottleneck approach. arXiv preprint arXiv:2205.06862  (2022)

\bibitem{adams2023fully}
Adams, J., Elhabian, S.: Fully bayesian vib-deepssm. arXiv preprint arXiv:2305.05797  (2023)

\bibitem{beg2005computing}
Beg, M.F., Miller, M.I., Trouv{\'e}, A., Younes, L.: Computing large deformation metric mappings via geodesic flows of diffeomorphisms. International journal of computer vision  \textbf{61}(2),  139--157 (2005)

\bibitem{bhalodia2021deepssm}
Bhalodia, R., Elhabian, S., Adams, J., Tao, W., Kavan, L., Whitaker, R.: Deepssm: A blueprint for image-to-shape deep learning models. arXiv preprint arXiv:2110.07152  (2021)

\bibitem{bhalodia2018deepssm}
Bhalodia, R., Elhabian, S.Y., Kavan, L., Whitaker, R.T.: Deepssm: a deep learning framework for statistical shape modeling from raw images. In: International Workshop on Shape in Medical Imaging. pp. 244--257. Springer (2018)

\bibitem{bhalodia2018deep}
Bhalodia, R., Goparaju, A., Sodergren, T., Morris, A., Kholmovski, E., Marrouche, N., Cates, J., Whitaker, R., Elhabian, S.: Deep learning for end-to-end atrial fibrillation recurrence estimation. In: 2018 Computing in Cardiology Conference (CinC). vol.~45, pp.~1--4. IEEE (2018)

\bibitem{biffi2020explainable}
Biffi, C., Cerrolaza, J.J., Tarroni, G., Bai, W., De~Marvao, A., Oktay, O., Ledig, C., Le~Folgoc, L., Kamnitsas, K., Doumou, G., et~al.: Explainable anatomical shape analysis through deep hierarchical generative models. IEEE transactions on medical imaging  \textbf{39}(6),  2088--2099 (2020)

\bibitem{cates2017shapeworks}
Cates, J., Elhabian, S., Whitaker, R.: Shapeworks: particle-based shape correspondence and visualization software. In: Statistical Shape and Deformation Analysis, pp. 257--298. Elsevier (2017)

\bibitem{cates2007shape}
Cates, J., Fletcher, P.T., Styner, M., Shenton, M., Whitaker, R.: Shape modeling and analysis with entropy-based particle systems. In: Biennial International Conference on Information Processing in Medical Imaging. pp. 333--345. Springer (2007)

\bibitem{davies2002minimum}
Davies, R.H., Twining, C.J., Cootes, T.F., Waterton, J.C., Taylor, C.J.: A minimum description length approach to statistical shape modeling. IEEE transactions on medical imaging  \textbf{21}(5),  525--537 (2002)

\bibitem{fayek2020progressive}
Fayek, H.M., Cavedon, L., Wu, H.R.: Progressive learning: A deep learning framework for continual learning. Neural Networks  \textbf{128},  345--357 (2020)

\bibitem{fuessinger2019virtual}
Fuessinger, M.A., Schwarz, S., Neubauer, J., Cornelius, C.P., Gass, M., Poxleitner, P., Zimmerer, R., Metzger, M.C., Schlager, S.: Virtual reconstruction of bilateral midfacial defects by using statistical shape modeling. Journal of Cranio-Maxillofacial Surgery  \textbf{47}(7),  1054--1059 (2019)

\bibitem{gao2016snr}
Gao, T., Du, J., Dai, L.R., Lee, C.H.: Snr-based progressive learning of deep neural network for speech enhancement. In: Interspeech. pp. 3713--3717 (2016)

\bibitem{gardner2013point}
Gardner, G., Morris, A., Higuchi, K., MacLeod, R., Cates, J.: A point-correspondence approach to describing the distribution of image features on anatomical surfaces, with application to atrial fibrillation. In: 2013 IEEE 10th International Symposium on Biomedical Imaging. pp. 226--229. IEEE (2013)

\bibitem{gerig2001shape}
Gerig, G., Styner, M., Jones, D., Weinberger, D., Lieberman, J.: Shape analysis of brain ventricles using spharm. In: Proceedings IEEE Workshop on Mathematical Methods in Biomedical Image Analysis (MMBIA 2001). pp. 171--178. IEEE (2001)

\bibitem{goparaju2022benchmarking}
Goparaju, A., Iyer, K., Bone, A., Hu, N., Henninger, H.B., Anderson, A.E., Durrleman, S., Jacxsens, M., Morris, A., Csecs, I., et~al.: Benchmarking off-the-shelf statistical shape modeling tools in clinical applications. Medical Image Analysis  \textbf{76},  102271 (2022)

\bibitem{harris2013statistical}
Harris, M.D., Datar, M., Whitaker, R.T., Jurrus, E.R., Peters, C.L., Anderson, A.E.: Statistical shape modeling of cam femoroacetabular impingement. Journal of Orthopaedic Research  \textbf{31}(10),  1620--1626 (2013)

\bibitem{karras2017progressive}
Karras, T., Aila, T., Laine, S., Lehtinen, J.: Progressive growing of gans for improved quality, stability, and variation. arXiv preprint arXiv:1710.10196  (2017)

\bibitem{li2018deep}
Li, C., Zia, M.Z., Tran, Q.H., Yu, X., Hager, G.D., Chandraker, M.: Deep supervision with intermediate concepts. IEEE transactions on pattern analysis and machine intelligence  \textbf{41}(8),  1828--1843 (2018)

\bibitem{liu2016learning}
Liu, Y., Lew, M.S.: Learning relaxed deep supervision for better edge detection. In: Proceedings of the IEEE conference on computer vision and pattern recognition. pp. 231--240 (2016)

\bibitem{loshchilov2016sgdr}
Loshchilov, I., Hutter, F.: Sgdr: Stochastic gradient descent with warm restarts. arXiv preprint arXiv:1608.03983  (2016)

\bibitem{park2019deepsdf}
Park, J.J., Florence, P., Straub, J., Newcombe, R., Lovegrove, S.: Deepsdf: Learning continuous signed distance functions for shape representation. In: Proceedings of the IEEE/CVF conference on computer vision and pattern recognition. pp. 165--174 (2019)

\bibitem{ronneberger2015u}
Ronneberger, O., Fischer, P., Brox, T.: U-net: Convolutional networks for biomedical image segmentation. In: Medical Image Computing and Computer-Assisted Intervention--MICCAI 2015: 18th International Conference, Munich, Germany, October 5-9, 2015, Proceedings, Part III 18. pp. 234--241. Springer (2015)

\bibitem{styner2006framework}
Styner, M., Oguz, I., Xu, S., Brechb{\"u}hler, C., Pantazis, D., Levitt, J.J., Shenton, M.E., Gerig, G.: Framework for the statistical shape analysis of brain structures using spharm-pdm. The insight journal (1071), ~242 (2006)

\bibitem{wang2015training}
Wang, L., Lee, C.Y., Tu, Z., Lazebnik, S.: Training deeper convolutional networks with deep supervision. arXiv preprint arXiv:1505.02496  (2015)

\bibitem{wu2019progressive}
Wu, Y., Lin, Y., Dong, X., Yan, Y., Bian, W., Yang, Y.: Progressive learning for person re-identification with one example. IEEE Transactions on Image Processing  \textbf{28}(6),  2872--2881 (2019)

\bibitem{zhang2018deep}
Zhang, Y., Chung, A.: Deep supervision with additional labels for retinal vessel segmentation task. In: International conference on medical image computing and computer-assisted intervention. pp. 83--91. Springer (2018)

\bibitem{zhao2008hippocampus}
Zhao, Z., Taylor, W.D., Styner, M., Steffens, D.C., Krishnan, K.R.R., MacFall, J.R.: Hippocampus shape analysis and late-life depression. PLoS One  \textbf{3}(3),  e1837 (2008)

\bibitem{zhou2021multi}
Zhou, Y., Chen, H., Li, Y., Liu, Q., Xu, X., Wang, S., Yap, P.T., Shen, D.: Multi-task learning for segmentation and classification of tumors in 3d automated breast ultrasound images. Medical Image Analysis  \textbf{70},  101918 (2021)

\end{thebibliography}

\end{document}